\documentclass[10pt, conference]{IEEEtran}

\IEEEoverridecommandlockouts
\usepackage{cite}
\usepackage{amsmath,amssymb,amsfonts}
\usepackage{algorithmic}
\usepackage{graphicx}
\usepackage{textcomp}
\usepackage{xcolor}
\usepackage{soul}
\usepackage{graphicx} 
\usepackage{array}    
\usepackage{booktabs} 
\usepackage{makecell}
\def\BibTeX{{\rm B\kern-.05em{\sc i\kern-.025em b}\kern-.08em
    T\kern-.1667em\lower.7ex\hbox{E}\kern-.125emX}}
\begin{document}

\title{FRIEND: Federated Learning for Joint Optimization of multi-RIS Configuration and Eavesdropper Intelligent Detection in B5G Networks}

\author{
    \IEEEauthorblockN{Maria Lamprini A. Bartsioka\IEEEauthorrefmark{1}, Ioannis A. Bartsiokas\IEEEauthorrefmark{1}, Anastasios K. Papazafeiropoulos\IEEEauthorrefmark{2}, \\Maria A. Seimeni\IEEEauthorrefmark{1}, Dimitra I. Kaklamani\IEEEauthorrefmark{1} and Iakovos S. Venieris\IEEEauthorrefmark{3}}
    \IEEEauthorblockA{\IEEEauthorrefmark{1}\textit{Microwave and Fiber Optics Laboratory, School of Electrical and Computer Engineering,} \\\textit{National Technical University of Athens, 9 Heroon Polytechneiou str, Zografou 15780, Athens, Greece}}
    \IEEEauthorblockA{\IEEEauthorrefmark{2}\textit{Communications and Intelligent Systems Research Group, University of Hertfordshire,} \\ \textit{AL10 9AB Hatfield, U.K.}}
    \IEEEauthorblockA{\IEEEauthorrefmark{3}\textit{Intelligent Communications and Broadband Networks Laboratory, School of Electrical and Computer Engineering,} \\ \textit{National Technical University of Athens, 9 Heroon Polytechneiou str, Zografou 15780, Athens, Greece}}
Emails: bartsiokamarilina@mail.ntua.gr, giannismpartsiokas@mail.ntua.gr, tapapazaf@gmail.com, \\ mseimeni@icbnet.ece.ntua.gr, dkaklam@mail.ntua.gr, venieris@cs.ntua.gr}

\maketitle

\begin{abstract}
As wireless systems evolve toward Beyond 5G (B5G), the adoption of cell-free (CF) millimeter-wave (mmWave) architectures combined with Reconfigurable Intelligent Surfaces (RIS) is emerging as a key enabler for ultra-reliable, high-capacity, scalable, and secure Industrial Internet of Things (IIoT) communications. However, safeguarding these complex and distributed environments against eavesdropping remains a critical challenge, particularly when conventional security mechanisms struggle to overcome scalability, and latency constraints. In this paper, a novel framework for detecting malicious users in RIS-enhanced cell-free mmWave networks using Federated Learning (FL) is presented. The envisioned setup features multiple access points (APs) operating without traditional cell boundaries, assisted by RIS nodes to dynamically shape the wireless propagation environment. Edge devices collaboratively train a Deep Convolutional Neural Network (DCNN) on locally observed Channel State Information (CSI), eliminating the need for raw data exchange. Moreover, an early-exit mechanism is incorporated in that model to jointly satisfy computational complexity requirements. Performance evaluation indicates that the integration of FL and multi-RIS coordination improves approximately 30\% the achieved secrecy rate (SR) compared to baseline non-RIS-assisted methods while maintaining near-optimal detection accuracy levels. This work establishes a distributed, privacy-preserving approach to physical layer eavesdropping detection tailored for next-generation IIoT deployments.
 
\end{abstract}


\section{Introduction}

The wireless communications landscape is undergoing a profound transformation, driven by the increasing demands for ultra-fast connectivity, pervasive sensing, and intelligent automation. While fifth-generation (5G) networks have delivered significant advancements in bandwidth, latency, and device density, the next wave of innovation, encompassing Beyond 5G (B5G) and the upcoming sixth generation (6G) systems, aims to integrate communication, computation, and intelligence at an unprecedented scale \cite{b1}. These future networks are expected to support emerging use cases such as autonomous systems, remote industrial control, extended reality (XR), and massive-scale Internet of Things (IoT) deployments. Key enabling technologies include millimeter-wave (mmWave) transmission, Reconfigurable Intelligent Surfaces (RIS), cell-free architectures, artificial intelligence (AI) at the edge, and distributed learning frameworks. However, as communication systems become more decentralized and heterogeneous, ensuring security and privacy across the physical layer becomes a major research and design challenge \cite{b2}.

In this context, Physical Layer Security (PLS) has emerged as a promising paradigm to complement traditional cryptographic approaches by leveraging the inherent randomness and physical characteristics of wireless channels \cite{b3}. Unlike encryption-based techniques that operate at higher Open System Interconnections (OSI) layers, PLS mechanisms provide security guarantees at the signal level (L1 and L2 of the OSI), enabling real-time authentication, secrecy, and resilience without incurring significant computational overhead. Methods such as secure beamforming, artificial noise injection, channel-based authentication, and anomaly detection based on Channel State Information (CSI) are considered as promising (proactive: preventive, active: detecting) against attacks such as spoofing, jamming and eavesdropping. However, as B5G systems grow increasingly in terms of complexity and decentralization, the design of PLS mechanisms becomes more challenging, especially in scenarios involving distributed processing, mobility, and high-dimensional channel variations.

One such architecture is cell-free massive Multiple-Input and Multiple-Output (CF-mMIMO) orientations, where a large number of distributed access points (APs) collaboratively serve users across a given area with dynamic irregular cellular boundaries. Unlike conventional cellular topologies, where cell boundaries are fixed regular formations, CF-mMIMO enhances spectral efficiency and link reliability by eliminating inter-cell interference (ICI) and enabling coherent joint transmission \cite{b4,new}. It is particularly well-suited for Industrial Internet of Things (IIoT) deployments, where reliable and low-latency connectivity is essential for automation and control. However, the spatial distribution of APs and the dynamic association of users present significant challenges for centralized security management, making distributed PLS techniques increasingly relevant \cite{b5}.

Complementing the CF-mMIMO architecture, mmWave communication has been introduced to address the bandwidth limitations of sub-6 GHz systems by exploiting frequency bands above 24 GHz. mmWave systems can achieve increased data rates and ultra-low latency, but are inherently susceptible to high path loss, sensitivity to blockage, and rapid channel fading \cite{b6, new2}. These characteristics impact both the performance and the security of the link, particularly in the presence of adversarial nodes \cite{b7}.

To further improve wireless propagation in high-frequency regimes, RIS have gained attention as a low-cost and energy-efficient solution. RISs are programmable meta-surfaces that can manipulate the phase, amplitude, and direction of incident electromagnetic waves. By deploying multiple RIS elements in the environment, it is possible to create favorable propagation conditions for legitimate users, while degrading the signal quality received by potential eavesdroppers \cite{b8, new3}. However, integrating RIS into security-aware communication systems introduces new design questions, such as joint beam optimization, attack surface reduction, and CSI acquisition under adversarial conditions \cite{b9}.

In the complex content of the coexistence of the aforementioned technologies in the modern wireless communications domain, machine learning (ML), and in particular deep learning (DL), has emerged as a powerful tool for enhancing PLS through data-driven adaptive signal processing and intelligent radio resource management (RRM) \cite{b10}. However, conventional ML approaches often rely on centralized training, which may not be practical in IIoT scenarios due to privacy concerns, limited bandwidth, and distributed data sources. To overcome these limitations, Federated Learning (FL) has been proposed as a decentralized learning paradigm, where local devices collaboratively train a global model without sharing raw data. This makes FL particularly suitable for next-generation wireless networks, enabling privacy-preserving intelligence at scale in edge topologies \cite{b11}.

In this work, the challenges mentioned above are addressed by proposing an intelligent FL-based framework for eavesdropper detection in RIS-assisted cell-free mmWave IIoT orientations. The framework leverages local CSI knowledge at edge nodes to train DL models capable of distinguishing between legitimate and malicious transmissions at the physical layer. To protect user privacy and reduce communication overhead, model training is performed in a federated manner, without exchanging raw data with central aggregation entities. Additionally, RIS units are incorporated into the simulated network topology to enhance signal controllability and detection robustness.

The remainder of this paper is structured as follows: Section II reviews relevant literature on PLS and ML-aided eavesdropper detection in mmWave RIS scenarios. Moreover the motivation of our proposed work is depicted in terms of overall system's engineering. Section III presents the system model and problem formulation. Section IV details the DL architectures and federated training scheme, while section V provides simulation results and performance evaluation. Finally, Section VI concludes the paper and outlines directions for future research.

\section{Concept Definition \& Requirements Analysis}

\subsection{Relevant Literature on the field}
Traditional PLS techniques have focused on leveraging signal processing methods to enhance confidentiality and resilience. Approaches such as artificial noise (AN) injection, secure beamforming, and secrecy rate optimization have been widely adopted to degrade eavesdropper reception without compromising legitimate user performance \cite{b12, b13}. In the context of RIS-assisted systems, several works have explored passive beam design to maximize the signal-to-interference-and-noise ratio (SINR) of legitimate links while minimizing information leakage \cite{b14}. Likewise, in CF-mMIMO environments, coordinated AP transmission and user-centric clustering have been proposed to improve secrecy capacity under pilot contamination and channel uncertainty \cite{b15}. Despite their effectiveness, these methods often rely on perfect or near-perfect CSI and cannot readily adapt to dynamic and intelligent adversaries.

The integration of ML into RIS-enhanced networks has led to a new class of security-aware designs. For example, in \cite{b16}, the authors explore the security challenges in UAV-assisted RIS networks. A DL framework based on Long Short-Term Memory Deep Deterministic Policy Gradient (LSTM-DDPG) is proposed to detect and mitigate malicious threats in dynamic environments. Simulation results show significant improvements over the baseline methods, demonstrating the potential of combining DL, RIS control, and aerial mobility to enhance the robustness of next-generation wireless systems. Other approaches, as the one in \cite{b17}, utilize deep reinforcement learning (DRL) techniques in joint active and passive beamforming scenarios. The optimazation is performed using a DDPG algorithm to maximize the sum secrecy rate of trusted devices while ensuring Quality of Service (QoS) for all users. Results demonstrate a 2–2.5× secrecy rate improvement over benchmark schemes, highlighting the potential of DRL-based optimization in RIS-assisted secure networks.

Cell-free networks present both opportunities and challenges for secure communication. On the one hand, their distributed nature can improve secrecy by increasing spatial diversity and reducing eavesdropping zones. On the other hand, the lack of centralized coordination complicates joint security policy enforcement. In \cite{b18}, the authors propose a joint AP selection and power allocation strategy to maximize the spectral efficiency (SE) of legitimate users while maintaining positive secrecy SE, using a low-complexity accelerated projected gradient (APG) algorithm. Results show a 62\% gain in SE compared to baseline schemes and improved secrecy performance even with multiple active adversaries. Meanwhile, our previous work in \cite{b19} evaluated models such as Random Forests (RF), Deep Convolutional Neural Networks (DCNNs), and Long Short-Term Memory (LSTM) networks, using CSI, location data, and transmission power as features to detect eavesdroppers. Results indicate that RF and DCNN models can achieve near-perfect detection accuracy with zero false alarms, highlighting the promise of AI-driven PLS in addressing emerging threats in next-generation wireless systems.

FL has gained traction as a privacy-preserving alternative to centralized ML in B5G/6G network orientations. For example, in \cite{b20}, the authors investigate a relay-assisted Federated Edge Learning (FEEL) system composed of multiple users, relays, and a central edge server, operating under latency and bandwidth constraints. A partial aggregation and spectrum multiplexing at the relays scheme is proposed, along with two bandwidth allocation strategies based on either instantaneous or statistical CSI. 

As reviewed above, numerous studies have explored the integration of PLS and ML/DL across diverse wireless scenarios, including RIS-assisted, cell-free, and mmWave propagation. However, most existing works examine these technologies disjointly, often neglecting the complex interplay between network topology, propagation characteristics, and distributed intelligence. The novelty of our work lies in the unified treatment of active eavesdropper detection within a cell-free mmWave network enhanced by multiple RIS units. By leveraging FL over CSI-based data representations, we combine these technologies into a comprehensive and scalable simulation framework that enables rigorous evaluation of detection performance, secrecy rate, and training efficiency in realistic B5G environments. The overall system is designed using both MATLAB R2025b (for problem formulation and dataset generation) and Python (for ML/DL/FL model training and evaluation).  

\subsection{System Requirements Definition}

The development of intelligent wireless B5G/6G systems -likewise their predecessors- are expected to ensure and further enhance their Availability, Reliability, Maintainability, Securability, Performance, Interoperability, and Scalability early in their conceptualization phase way forward to their design, integration, verification, validation and maintenance.

As already stated in previous Sections, in this study, our main focus falls on the Securability and Scalability merit of the system (Priority 1) without diminishing the significance of the rest pillars (Priority 2), though. In this context (Concept Definition Phase), Mission Analysis and Stakeholder Needs Definition were performed. Typically during this phase \cite{new4, new5}, the goals, the capabilities and measures of effectiveness (MOE) of the System of Interest (SoI) are identified along with the stakeholders feedback including degrees of freedom (e.g. standards, regulations, budget, existing technologies, existing (sub-)systems, risks, key drivers, maturity level, etc. It is reminded that the effort of this paper primarily focus on the study and evaluation of the FL Model (Priority 1).

To reproduce similar scenarios and jointly ensure the above mentioned system merits, the authors choose to develop the System-as-a-code; hence the eavesdropping-resilient CF-mMIMO RIS-assisted B5G/6G system simulator has been created. During the System Requirements Definition phase, the authors aim to technically translate the outputs into a Simulator Requirements List (presented in Table \ref{tab:req_analysis}), which shall serve as the major source of information for the technical processes (e.g Design Definition, Architecture Definition, Verification, Maintenance) of the following Sections. 

Specifically, the requirements should clearly communicate “what” (i.e. the functions) the simulator must do to satisfy stakeholders’ needs; hence the Model Based Systems Engineering Paradigm (MBSE) is followed. In our study, we organize the requirements into the following categories: Non-Function, Function (Performance), Fit (Operational), Form, Compliance. Table I summarizes a subset of representative requirements identified during the conceptual phase of the proposed scheme for eavesdropping detection in CF-mMIMO RIS-assisted B5G/6G systems. Each requirement is classified according to its type and includes a concise justification to support verification planning.

In practice, these requirements guide both architectural decomposition and subsystem interaction modeling. Ultimately, by applying a disciplined systems engineering framework, the proposed design process ensures that the non-functional requirements are not treated as afterthoughts but are instead embedded from the earliest stages of development.

\begin{table*}[!t]
\centering
\caption{Simulator Requirements}
\label{tab:req_analysis}
\resizebox{\textwidth}{!}{
\begin{tabular}{|p{0.4cm}|p{6cm}|p{2.2cm}|p{2cm}|p{2.5cm}|p{2cm}|}
\hline
\textbf{ID} & \textbf{Requirement Description} & \textbf{Category} & \textbf{Target \newline (Sub-)System} & \textbf{Challenge} & \textbf{Work} \\
\hline
R1 & The SoI shall support multiscenario channel data generation reflecting the 5V data paradigm (Volume, Variety, Velocity, Veracity, Value). & Fit (Operational) & MATLAB & SV Data & Current Work \\
\hline
R2 & The SoI shall provide reliable preprocessing and feature extraction to ensure balanced and representative datasets. & Fit (Operational) & MATLAB & SV Data & Current Work \\
\hline
R3 & The SoI shall incorporate bias–variance control mechanisms (e.g., regularization, cross-validation, early stopping) to guarantee model generalization. & Function (Performance) & PYTHON & Bias–Variance Balance & Current Work \\
\hline
R4 & The SoI shall jointly evaluate performance using ML accuracy metrics and communication QoS metrics. & Function (Performance) & MATLAB, PYTHON & Design Priorities & Current Work \\
\hline
R5 & The SoI shall integrate data generation and DL/ML decision in real-time to guarantee seamless simulation runs. & Fit (Operational) & MATLAB, PYTHON & MATLAB/PYTHON API & Future Work \\
\hline
R6 & The SoI shall integrate data generation and DL/ML in a reproducible manner to maintain interoperability and traceability of experiments. & Non-Functional & MATLAB, PYTHON & MATLAB/PYTHON API & Future Work \\
\hline
R7 & The SoI shall enable future upgrades for increased RIS configuration diversity and scalable B5G/6G architecture. & Function (Performance) & MATLAB & Current MATLAB functionality & Future Work \\
\hline
R8 & The SoI shall support real data loading to complement the evaluation of the proposed scheme. & Fit (Operational) & PYTHON or Unified MATLAB/PYTHON & Current MATLAB functionality & Future Work \\
\hline
R9 & The SoI shall guarantee availability, reliability, performance, scalability, and maintainability. & Non-Functional & PYTHON or Unified MATLAB/PYTHON & Trade-off between Non-Functional Requirements & Current Work \\
\hline
R10 & The SoI shall detect and prevent privacy leakages. & Compliance & PYTHON or Unified MATLAB/PYTHON & RIS Profile, Wireless Propagation Environment & Current Work \\
\hline
\end{tabular}
}
\end{table*}

\section{System Model}

\subsection{Overall System Architecture}\label{AA}

\begin{figure}[htbp]
\centerline{\includegraphics[width=1\linewidth]{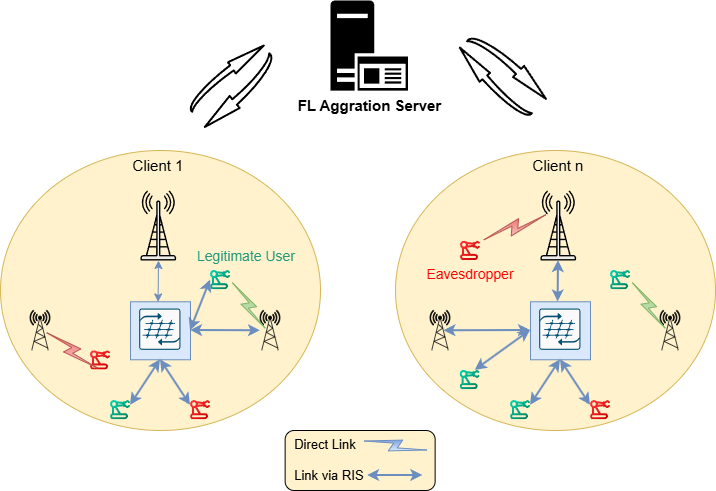}}
\caption{Cell-Free network topology.}
\label{fig1}
\end{figure}

As presented in Figure \ref{fig1} we consider a CF RIS-assisted mmWave network architecture, designed to emulate a dense industrial B5G environment. The scenario reflects a typical IIoT setting, where signal propagation is impacted by common obstacles such as industrial equipment, walls, and metallic surfaces. The system configuration adheres to the 3GPP specifications outlined in TR 38.901 \cite{b21} and TR 38.843 \cite{b22}.

The B5G/6G network topology comprises several distributed mmWave APs operating without fixed cell boundaries, multiple RIS strategically deployed throughout the area, a group of legitimate User Equipments (UEs), and a number of adversarial users (eavesdroppers). All entities are positioned in a 3D coordinate system that reflects an IIoT deployment. 

The following sets of entities represent the main components of the system:
\begin{itemize}
    \item $\mathcal{A} = \{a_1, a_2, ..., a_{N_{AP}}\}$, the set of APs.
    \item $\mathcal{R} = \{r_1, r_2, ..., r_{N_{RIS}}\}$, the set of RIS.
    \item $\mathcal{L} = \{\ell_1, \ell_2, ..., \ell_{N_{LU}}\}$, the set of legitimate UEs.
    \item $\mathcal{E} = \{e_1, e_2, ..., e_{N_E}\}$,  the set of eavesdroppers
\end{itemize}
where the total number of APs, RIS, legitimate UEs and eavesdroppers are denoted as \(N_{AP},N_{RIS},N_{LU}\) and \(N_E\) respectively.

The placement of both legitimate users and eavesdroppers follows the procedure in Section III of \cite{b19}. RIS units are placed in coverage-critical zones-so that they serve the entire topology equally-with the ability to manipulate incoming signals by tuning their reflection coefficients. This setup helps direct power towards intended users and weakens unintended leakage toward adversaries. The effective channel between APs and UEs (or eavesdroppers) includes both direct and RIS-reflected components.

Communication links are categorized as:
\begin{itemize}
    \item AP-to-UE links: $L_{a,\ell}$.
    \item AP-to-eavesdropper links: $L_{a,e}$.
    \item AP-to-RIS links: $L_{a,r}$.
    \item RIS-to-UE links: $L_{r,\ell}$
    \item RIS-to-eavesdropper links: $L_{r,e}$
\end{itemize}

\subsection{Problem Formulation}
Following the deployment of the RIS-enabled, CF mmWave network topology, the subsequent step involves formulating a rigorous communication model and defining performance metrics for PLS. The network operates under a Time Division Duplex (TDD) protocol, where each legitimate UE transmits Sounding Reference Signals (SRSs) to nearby APs for uplink channel estimation. These SRSs, based on OFDM waveforms, traverse the wireless medium and are subject to distortion due to multi-path fading, filtering effects, and additive white Gaussian noise (AWGN). However, their key role is assisting each AP in obtaining the mandatory CSI, in order to differentiate legitimate users from possible attackers. In this work, communication occurs through both direct AP–UE links and indirect RIS-assisted reflections. Each AP receives SRSs from the users and estimates the aggregated channel composed of the direct and reflected components. 

Specifically, for an AP $a\in\mathcal{A}$, a legitimate user $l\in\mathcal{L}$, and a RIS $r\in\mathcal{R}$, the overall uplink channel $h_{eff,a,l}$ is modeled as:
\[
\mathbf{h}_{eff,a,l} = \mathbf{h}_{dir,a,l} + \mathbf{G}^H_{r,b}\mathbf{\Theta}_{r}\mathbf{h}_{l,r} \tag{1}
\]

Where, $\mathbf{h}_{dir,a,l}\in\mathbb{C}^{N_r \times N_t}$ is the direct mmWave channel from legitimate UE $l$ and AP $a$, $\mathbf{h}_{l,r}\in\mathbb{C}$ is the channel from legitimate UE $l$ to RIS $r$, $\mathbf{G}^H_{r,b}\in\mathbb{C}^{N \times N_t}$ is the channel from RIS $r$ to AP $a$, and $\mathbf{\Theta}_{r}=diag(e^{j\theta_1},e^{j\theta_2},...,e^{j\theta_N})$ is the RIS reflection matrix, comprising $N$ phase shift elements. Finally, $N_t$ is the number of antenna elements in each AP, while single-antenna legitimate/malicious UE are assumed. A similar representation as the one in (1) can be used to model the effective channel for the malicious UEs (eavesdroppers), Moreover, downlink channels follow the same logic. This model enables each AP to capture both direct and intelligently reflected signal components.

To secure communication, the secrecy rate (SR) between legitimate UE $l$ and AP $a$ in the presence of a potential eavesdropper $e$, is defined as \cite{b19}:
\begin{gather*}
SR_{l,a} = [ C_{l,i,a} - C_{e,j,a}]^+ \\
= log_2(1+SINR_{l,i,a}) - log_2(1+SINR_{e,j,a}) \tag{2}
\end{gather*}
where \(C_{l,i,a}\) and \( C_{e,j,a}\) represent the respective channel capacities between the AP and the legitimate user (channel $i$) and an eavesdropper \(e \in \mathcal{E}\) (channel $j$). In the same way, \(SINR_{l,i,a}\) and \(SINR_{e,j,a}\) stand for the SINR of the corresponding channels including both direct and RIS-aided links as depicted in (1). The operator \([.]^+\) ensures non-negative SRs by taking the maximum between the computed value and zero.

To evaluate the overall system’s ability to maintain secure communication, and to compare the effectiveness of the proposed approaches in terms of both secrecy and data throughput, the \textit{average secrecy rate (ASR)} across all legitimate users and channels, is computed as \cite{b19}:
\begin{equation}
\overline{ASR} = \frac{1}{N_{AP}} \sum_{a=1}^{N_{AP}} \sum_{i=1}^{N_{LU}} SR_{l_{i,a}}
\tag{3}
\end{equation}

\section{FL-Based Eavesdropper Detection Mechanism}

\subsection{Data Gathering}
As highlighted in \cite{b1}, an essential step in building robust ML/DL models is the training phase, which heavily relies on precise, validated, and realistic datasets. However, real-world communication datasets are often difficult to access due to privacy constraints and proprietary restrictions. To overcome these limitations, we constructed a comprehensive simulation framework in MATLAB, to model an advanced IIoT deployment scenario within a cell-free mmWave B5G/6G environment.

The simulated network includes a distributed topology of APs, UEs, eavesdroppers, and multi-RIS. Key wireless parameters such as multi-path fading, blockage effects, large- and small-scale fading, SE, and RIS reflections are carefully integrated to emulate realistic channel conditions. As depicted in the previous Section OFDM-based SRS are transmitted from UEs to APs, and both direct and RIS-assisted paths are modeled. Each UE-AP RIS-aided link is represented through a complex-valued channel matrix, which is subsequently transformed into a CSI image. 

These high-dimensional representations capture both direct and reflected signal components, providing a rich dataset that supports accurate classification. Each sample is labeled as either a legitimate user (label 0) or an eavesdropper (label 1), with auxiliary metadata such as user position, transmit power, and serving node information included. To enhance realism, UEs are dynamically associated with the AP and RIS node offering the best received SNR, rather than based solely on distance.

\subsection{FL Scheme for Eavesdropper Detection}

To enable efficient and privacy-preserving eavesdropper detection in the proposed RIS-assisted CF mmWave network, an FL-based scheme that distributes model training across multiple APs is adopted. This means that the data collection process described above does not take place centrally, but at specific APs. In more detail, the APs that also function as FL clients are selected based on their location (so that they are at central points of the network) and on their distance from the RIS components (so that they can communicate with all of them for maximum signal amplification). Each one of those APs locally trains a DL model using its own dataset partition—composed of CSI samples and associated labels—without transmitting any raw data to the central server (hosted at the central AP of the whole topology). The server coordinates training rounds and periodically aggregates local model updates via the standard Federated Averaging (FedAvg) algorithm \cite{b23}, thereby maintaining both data privacy and training scalability.

\begin{figure}[htbp]
\centerline{\includegraphics[width=1\linewidth]{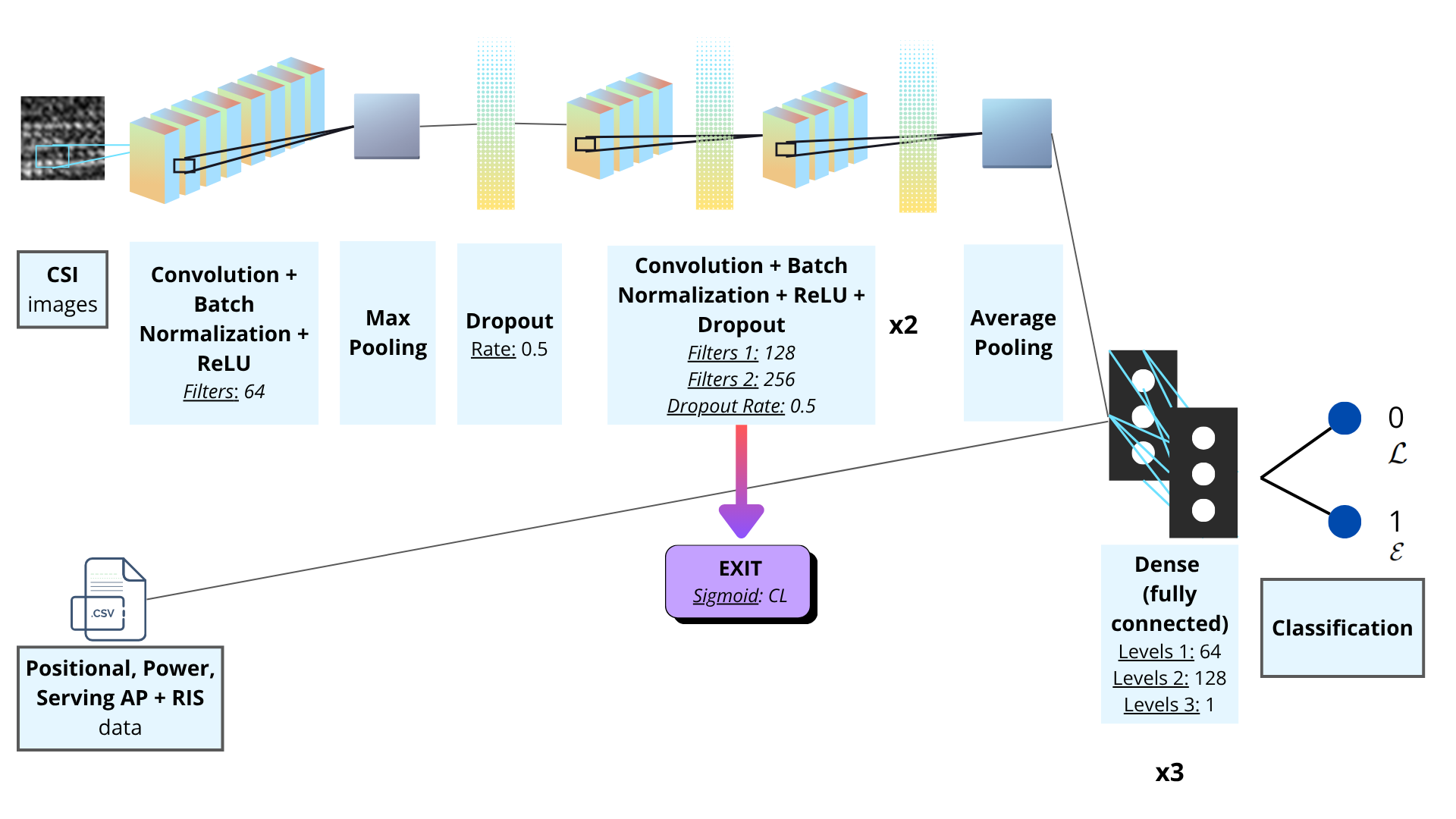}}
\caption{Proposed DCNN's Structure}
\label{fig2}
\end{figure}

In this framework, a Deep Convolutional Neural Network (DCNN) architecture is proposed due to its superior performance in processing complex CSI as depicted in \cite{b19}. The proposed model is a streamlined DCNN that ingests CSI images as input, extended by the fusion of additional information (UE transmission power and geolocation parameters (coordinates) of UE, serving AP and RIS nodes). The input tensor is expanded to accommodate both image-based and numerical features, with the first three channels corresponding to CSI maps and the next six capturing the supplementary metadata. The model was developed using the Keras Sequential API and optimized with the \textit{binary crossentropy} loss function. Its architecture is outlined in Figure \ref{fig2}.

Until this point, the systems requirements associated with R1-R4, R6, and R9 are already fulfilled by the MATLAB simulation and the proposed FL framework. To further enhance the adaptability of the DCNN, an early-exit mechanism was additionally integrated into the network architecture. The modified design introduces an auxiliary classifier after the second convolutional block, enabling the model to terminate inference early when the confidence level (CL) of the intermediate output exceeds a predefined threshold. During both local and global FL rounds, each client applies the same confidence policy, ensuring consistent inference behavior across distributed nodes. Therefore, the resource-constrained APs become capable of achieving reduced computational load and training time, while maintaining satisfactory classification performance and preserving privacy through the standard FedAvg aggregation process.

To assess model effectiveness, hyperparameters were tuned using grid search and cross-validation across multiple FL rounds. Afterwards, in the FedAvg aggregation scheme \cite{b23}, each AP-client independently trains the proposed DCNN on its local CSI-based data and periodically shares model weights—not raw samples—with a central aggregator (located at the central AP of the topology). Throughout the training process, synchronization between clients ensures that the global model captures diverse channel conditions and potential adversarial behaviors across the network. 

\section{Performance Evaluation}

\subsection{Analysis of FL Training Results and Convergence}

To validate the effectiveness of the proposed FL-based eavesdropper detection framework, extensive simulations were conducted under a configuration of a B5G/6G RIS-assisted CF-mMIMO mmWave network orientation. The simulation environment comprises 500 UEs, distributed randomly over the IIoT deployment area, with a fixed ratio of 70\% legitimate users and 30\% eavesdroppers. The infrastructure consists of 18 APs and 3 RIS units, strategically placed to enhance coverage and security performance. All entities operate at 28 GHz carrier frequency, and follow the specifications presented in Table \ref{tab:my_table}.
 
\begin{table}
    \centering
    \caption{Simulation Parameters}
    \label{tab:my_table}
    \begin{tabular}{|c|c|} \hline 
         \textbf{Parameter}& \textbf{Value/Assumption}\\ \hline 
         Frequency& 28 GHz (FR2)\\ \hline 
         Antennas' Height& APs: 8m, UEs: 1.5m\\ \hline 
         UEs Mobility& 3 km/h\\ \hline 
         Noise& 5 dB\\ \hline 
         Transmit Power& APs: 40 dB, LUs: 23 dB,
E: $>$23 dB\\ \hline 
         Subcarrier Spacing& 120 kHz\\ \hline 
 Bandwidth&400 MHz\\ \hline 
         Channel Modulation Symbols& 14\\ \hline 
         SRS Modulation Symbols& 12\\ \hline 
         Resource Blocks per Transmission& 60\\ \hline
         Number of Antennas& APs: 32, UEs: 1\\ \hline
         RIS phase dimensions& 10x20\\ \hline
         Number of FL-clients& 3\\ \hline
         Early-exit Confidence Levels & 55\% and 70\%\\ \hline
    \end{tabular}
\end{table}

\begin{figure}[htbp]
\centerline{\includegraphics[width=0.8\linewidth]{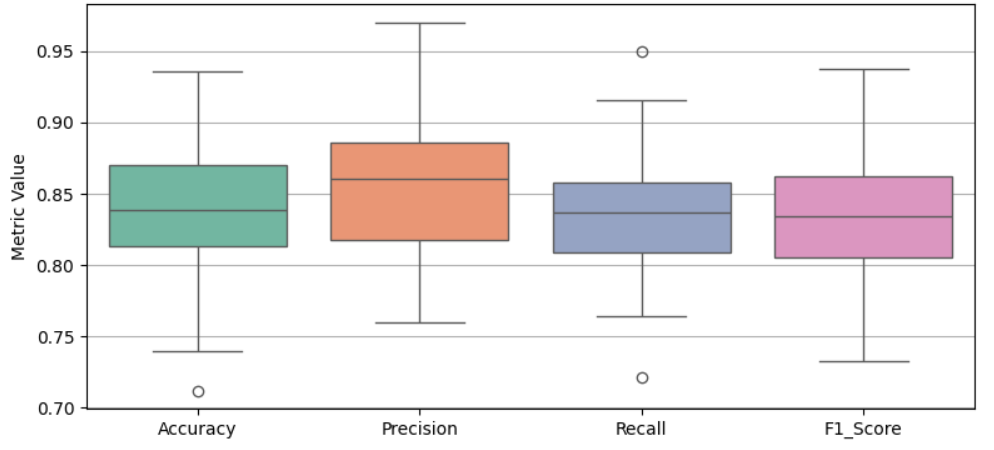}}
\caption{Distribution of DCNN Performance Metrics}
\label{fig3}
\end{figure}

By utilizing the parameters outlined previously, the MATLAB B5G system and link-level simulator generates the dataset that serves as input for the suggested DL model. During the training phase of all clients, an 80\%-20\% split between training and testing sets has been utilized. The challenge of detecting eavesdroppers is analyzed as a classification problem. We perform 100 Monte-Carlo simulations, each corresponding to different RIS phase shifts. The performance of the proposed DCNN model was tested with the use of Python library TensorFlow Federated across all the generated datasets. Figure \ref{fig3} illustrates the distribution of the known classification metrics obtained during the evaluation phase and concerning both classes. The metrics considered include Accuracy, Precision, Recall, and F1-score.

As observed, the accuracy values range from 0.71 to 0.93, with the interquartile range centered around 0.84. The recall metric, which reflects the model’s ability to detect true instances, shows slightly higher variability, reaching up to 0.95, while precision remains consistently high. Importantly, the F1-score, a balanced metric combining precision and recall, remains robust across all validation rounds, indicating that the model avoids both false alarms and missed detections.

The variability in recall stems from the high similarity between some eavesdropper and legitimate CSI samples, especially in edge cases where UEs are positioned near RIS units and exhibit low power discrepancies. Nonetheless, the model maintains high overall generalization performance despite the data heterogeneity introduced by the FL paradigm. The low number of false negatives (as reflected by high recall) is particularly critical in PLS scenarios, where failing to detect a malicious user can directly compromise communication secrecy.

\begin{figure}[htbp]
\centerline{\includegraphics[width=0.8\linewidth]{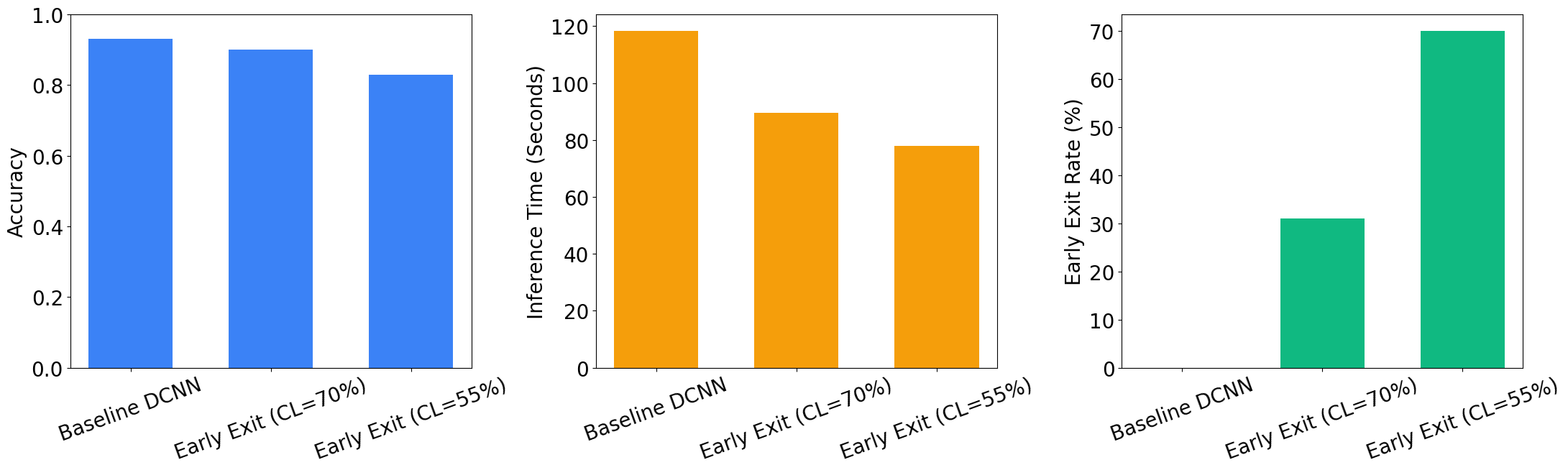}}
\caption{Comparison of Different implementation of DCNN model}
\label{fig4}
\end{figure}

At this point, it is important to evaluate the proposed scheme for B5G/6G systems in terms of the demand for intelligent mechanisms to jointly satisfy various requirements. For this reason, the early-exit extension of the DCNN model was investigated, aiming to assess the computational efficiency in a way that aligns with the bias–variance and latency requirements (R3,R5) defined in Section II-B. The findings are depicted in Figure \ref{fig4}. For the purpose of this figure, we rely exclusively on the best-performing RIS phase, as far as accuracy is concerned. It is obvious that by introducing an early-exit point after the second convolutional block, the inference time is substantially reduced up to 35\% for CL = 70\% and 45\% for CL = 55\%—while maintaining competitive accuracy levels ($>$ 0.83). The corresponding early-exit rates confirm that a significant proportion of samples can be confidently classified before full model traversal, offering notable latency gains without a severe accuracy drop. In other words, the early-exit mechanism emerges as a promising solution for minimizing the trade-off between a lightweight and adaptive, yet effective and reliable, FL model.

\subsection{Analysis of Communication Performance and Confidentiality}

\begin{figure}[htbp]
\centerline{\includegraphics[width=0.8\linewidth]{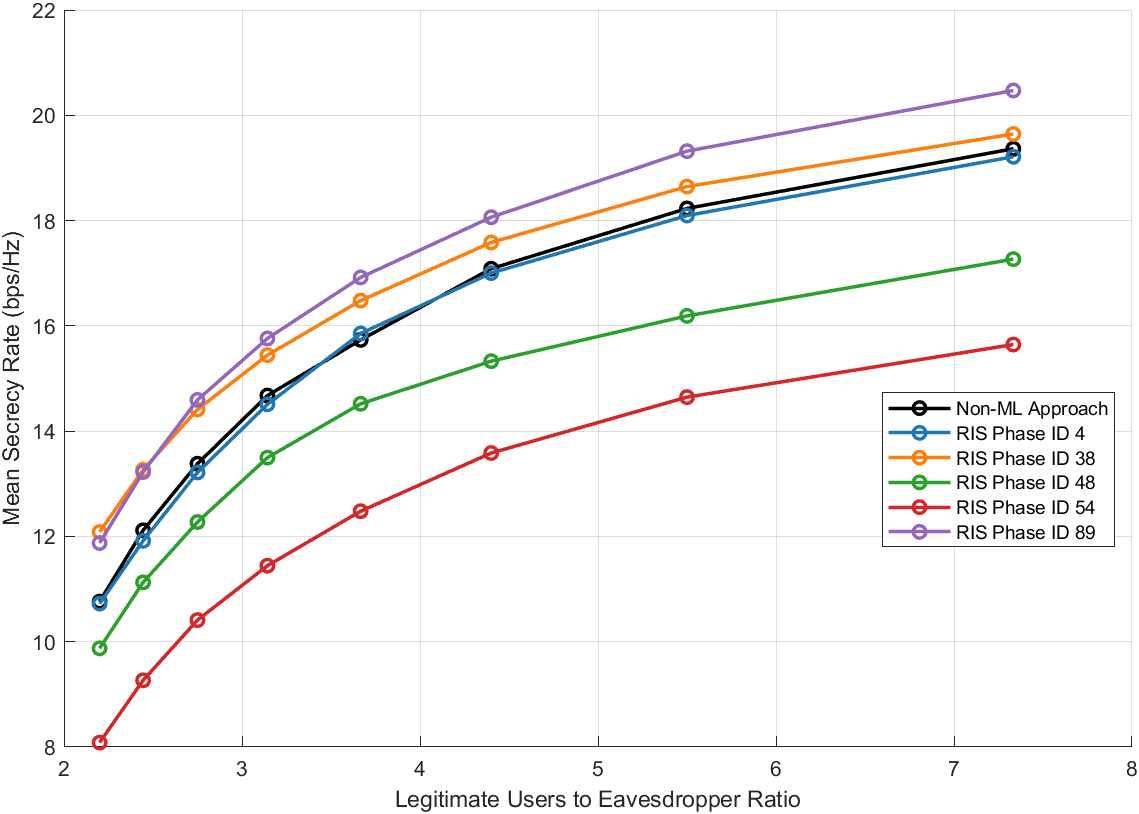}}
\caption{Average FL-based System's Secrecy Rate across different RIS configurations}
\label{fig5}
\end{figure}

As an additional step, to assess how RIS configurations affect PLS, a new evaluation round of the computed ASR, based on (3) was executed. The five best-performing RIS phase settings -based on the accuracy metric- are evaluated across a range of different $\mathcal{L}-to-\mathcal{E}$ ratios. The results are depicted in Figure \ref{fig5}.

It is evident that for all RIS phase configurations, the mean SR improves as the $\mathcal{L}-to-\mathcal{E}$ ratio increases. This behavior is expected, as a higher density of legitimate users statistically reduces the exposure of APs to malicious links. However, the degree of improvement is highly dependent on the RIS phase matrix used.

The configuration with RIS Phase ID 89 consistently achieves the highest SR, exceeding 20 bps/Hz in high $\mathcal{L}-to-\mathcal{E}$ ratios. At first sight, this indicates that this specific phase design offers highly constructive reflection paths for legitimate links while simultaneously inducing destructive interference at eavesdropper locations. In contrast, Phase ID 54 leads to significantly reduced secrecy rates, implying suboptimal signal redirection that may inadvertently boost eavesdropper reception.

However, at this point, one has to consider the above in conjunction to the "Non-ML Approach" SR curve, because higher SRs should not always be interpreted as superiority. The non-ML curve is computed based on the real classification labels and it can be observed that it closely matches the one of RIS Phase ID 4. This alignment validates that the classifier captures meaningful physical layer patterns and provides a realistic view of the network’s secrecy condition. Interestingly, some other Phase IDs outperform these values, but this stem from biased RIS configurations that inherently favor legitimate users, regardless of real-time adversary behavior. In other words, these phase shifts lead to lower values in recall for the eavesdropper class, even if the overall accuracy remains high.

To conclude, it is significant to evaluate the proposed FL-based approach against centralized non-RIS-aided state-of-the-art ones. For this purpose, the ASR performance can be compared to the one presented in Figure 5 of \cite{b19} for the $\mathcal{L}-to-\mathcal{E}$ ratio range between 2 and 5.5. From this comparative study, it is observed that the best-performing RIS phase (ID 4) achieves ~30\% greater ASR compared to the best-performing approach of \cite{b19}.

\section{Conclusions}

This work followed the Systems Engineering processes to define System Requirements and proposed a novel framework for intelligent eavesdropper detection in RIS-assisted, CF-mMIMO mmWave networks tailored to dense heterogeneous industrial B5G environments. The system leveraged the benefits of FL to collaboratively train a DCNN model across distributed APs, without exchanging raw CSI. The integration of RIS offered a joint optimization in this problem, by enabling both remarkable classification accuracy and suitable configuration for enhanced security over the wireless channel. The findings underline two key findings. Firstly, that intelligent classifiers trained under FL can reliably detect malicious transmissions, with accuracy and precision approaching 95\%. Secondly, that optimal RIS configurations can substantially improve the system’s PLS secreacy rate up to 18 bps/Hz when correctly tuned. Future directions of this research include the evaluation of the proposed scheme in real data (to also fulfill R8). Another open issue is the evaluation of the model in alternative or scalable RIS configurations (in line with R7 urge). Finally, the early-exit mechanism could be also investigated under offloading tasks.


\begin{thebibliography}{00}
\bibitem{b1} I. A. Bartsiokas, P. K. Gkonis, D. I. Kaklamani and I. S. Venieris, ``ML-Based Radio Resource Management in 5G and Beyond Networks: A Survey,'' \textit{IEEE Access}, vol. 10, pp. 83507-83528, 2022, doi: 10.1109/ACCESS.2022.3196657.
\bibitem{b2} A. Giannopoulos, A., P. K. Gkonis, A. Kalafatelis, N. Nomikos, S. Spantideas, P. Trakadas, T. Syriopoulos, ``From 6G to SeaX-G: Integrated 6G TN/NTN for AI-Assisted Maritime Communications—Architecture, Enablers, and Optimization Problems,'' \textit{J. Mar. Sci. Eng.} vol. 13, pp. 1103, 2025, https://doi.org/10.3390/jmse13061103
\bibitem{b3} L. Mucchi et al., ``Physical-Layer Security in 6G Networks,'' in \textit{IEEE Open Journal of the Communications Society}, vol. 2, pp. 1901-1914, 2021, doi: 10.1109/OJCOMS.2021.3103735.
\bibitem{b4} H. Q. Ngo, G. Interdonato, E. G. Larsson, G. Caire and J. G. Andrews, ``Ultradense Cell-Free Massive MIMO for 6G: Technical Overview and Open Questions,'' in \textit{Proceedings of the IEEE}, vol. 112, no. 7, pp. 805-831, July 2024, doi: 10.1109/JPROC.2024.3393514.
\bibitem{new} A. Papazafeiropoulos, E. Björnson, P. Kourtessis, S. Chatzinotas and J. M. Senior, ``Scalable Cell-Free Massive MIMO Systems: Impact of Hardware Impairments,'' \textit{IEEE Transactions on Vehicular Technology}, vol. 70, no. 10, pp. 9701-9715, Oct. 2021, doi: 10.1109/TVT.2021.3109341.
\bibitem{b5} H. He, X. Yu, J. Zhang, S. Song and K. B. Letaief, ``Cell-Free Massive MIMO for 6G Wireless Communication Networks,'' in \textit{Journal of Communications and Information Networks}, vol. 6, no. 4, pp. 321-335, Dec. 2021, doi: 10.23919/JCIN.2021.9663100.
\bibitem{b6} W. Deng, M. Li, M. -M. Zhao, M. -J. Zhao and O. Simeone, ``CSI Transfer From Sub-6G to mmWave: Reduced-Overhead Multi-User Hybrid Beamforming,'' in \textit{IEEE Journal on Selected Areas in Communications}, vol. 43, no. 3, pp. 973-987, March 2025, doi: 10.1109/JSAC.2025.3536539. 
\bibitem{new2} E. Shi et al., ``RIS-Aided Cell-Free Massive MIMO Systems for 6G: Fundamentals, System Design, and Applications,'' \textit{Proceedings of the IEEE}, vol. 112, no. 4, pp. 331-364, April 2024, doi: 10.1109/JPROC.2024.3404491. 
\bibitem{b7} S. Sobhi-Givi, M. Nouri, M. G. Shayesteh, H. Behroozi, H. H. Kwon and M. J. Piran, ``Efficient Optimization in RIS-Assisted UAV System Using Deep Reinforcement Learning for mmWave-NOMA 6G Communications,'' in \textit{IEEE Internet of Things Journal}, vol. 12, no. 14, pp. 26042-26057, 15 July15, 2025, doi: 10.1109/JIOT.2025.3553176.
\bibitem{b8} A. Papazafeiropoulos, J. An, P. Kourtessis, T. Ratnarajah and S. Chatzinotas, ``Achievable Rate Optimization for Stacked Intelligent Metasurface-Assisted Holographic MIMO Communications,'' in \text{IEEE Transactions on Wireless Communications}, vol. 23, no. 10, pp. 13173-13186, Oct. 2024, doi: 10.1109/TWC.2024.3399318.
\bibitem{new3} A. Papazafeiropoulos, C. Pan, P. Kourtessis, S. Chatzinotas and J. M. Senior, ``Intelligent Reflecting Surface-Assisted MU-MISO Systems With Imperfect Hardware: Channel Estimation and Beamforming Design,''  \textit{IEEE Transactions on Wireless Communications}, vol. 21, no. 3, pp. 2077-2092, March 2022, doi: 10.1109/TWC.2021.3109391.
\bibitem{b9} C. Pan et al., ``Reconfigurable Intelligent Surfaces for 6G Systems: Principles, Applications, and Research Directions,'' in \textit{IEEE Communications Magazine}, vol. 59, no. 6, pp. 14-20, June 2021, doi: 10.1109/MCOM.001.2001076.
\bibitem{b10} R. Sun et al., ``A Comprehensive Survey of Knowledge-Driven Deep Learning for Intelligent Wireless Network Optimization in 6G,'' in \textit{IEEE Communications Surveys \& Tutorials}, doi: 10.1109/COMST.2025.3574765. 
\bibitem{b11} Q. Duan, J. Huang, S. Hu, R. Deng, Z. Lu and S. Yu, ``Combining Federated Learning and Edge Computing Toward Ubiquitous Intelligence in 6G Network: Challenges, Recent Advances, and Future Directions,'' in \textit{IEEE Communications Surveys\& Tutorials}, vol. 25, no. 4, pp. 2892-2950, Fourthquarter 2023, doi: 10.1109/COMST.2023.3316615.
\bibitem{b12} W. Khalid, M. A. U. Rehman, T. Van Chien, Z. Kaleem, H. Lee and H. Yu, ``Reconfigurable Intelligent Surface for Physical Layer Security in 6G-IoT: Designs, Issues, and Advances,'' in \textit{IEEE Internet of Things Journal}, vol. 11, no. 2, pp. 3599-3613, 15 Jan.15, 2024, doi: 10.1109/JIOT.2023.3297241. 
\bibitem{b13} G. Jang, D. Kim, I. -H. Lee and H. Jung, ``Cooperative Beamforming With Artificial Noise Injection for Physical-Layer Security,'' in \textit{IEEE Access}, vol. 11, pp. 22553-22573, 2023, doi: 10.1109/ACCESS.2023.3252503
\bibitem{b14} 3GPP, G. C. Alexandropoulos, K. D. Katsanos, M. Wen and D. B. Da Costa, ``Counteracting Eavesdropper Attacks Through Reconfigurable Intelligent Surfaces: A New Threat Model and Secrecy Rate Optimization,'' in \textit{IEEE Open Journal of the Communications Society}, vol. 4, pp. 1285-1302, 2023, doi: 10.1109/OJCOMS.2023.3282814. 
\bibitem{b15} W. Li, N. Wang, L. Jiao and K. Zeng, ``Physical Layer Spoofing Attack Detection in MmWave Massive MIMO 5G Networks,'' in \textit{IEEE Access}, vol. 9, pp. 60419-60432, 2021, doi: 10.1109/ACCESS.2021.3073115.
\bibitem{b16} U. A. Mughal, Y. Alkhrijah, A. Almadhor and C. Yuen, ``Deep Learning for Secure UAV-Assisted RIS Communication Networks,'' in \textit{IEEE Internet of Things Magazine}, vol. 7, no. 2, pp. 38-44, March 2024, doi: 10.1109/IOTM.001.2300132.
\bibitem{b17} R. Saleem, W. Ni, M. Ikram and A. Jamalipour, ``Deep-Reinforcement-Learning-Driven Secrecy Design for Intelligent-Reflecting-Surface-Based 6G-IoT Networks,'' in \textit{IEEE Internet of Things Journal}, vol. 10, no. 10, pp. 8812-8824, 15 May15, 2023, doi: 10.1109/JIOT.2022.3232360.
\bibitem{b18} Y. S. Atiya, Z. Mobini, H. Q. Ngo and M. Matthaiou, ``Cell-Free Massive MIMO With Multiple Active Eavesdroppers,'' in \textit{IEEE Open Journal of the Communications Society}, vol. 6, pp. 1859-1872, 2025, doi: 10.1109/OJCOMS.2025.3534640.
\bibitem{b19}  M.-L. Bartsioka, I. Bartsiokas, P. Gkonis, D. Kaklamani, and I. Venieris, ``ML-enabled eavesdropper detection in beyond 5G IIoT networks,`` 2025 IEEE Symposium on Computers and Communications (ISCC), Bologna, Italy, 2025, pp 1-6, accepted, Available online: https://arxiv.org/abs/2505.07837.
\bibitem{b20} L. Chen, L. Fan, X. Lei, T. Q. Duong, A. Nallanathan and G. K. Karagiannidis, ``Relay-Assisted Federated Edge Learning: Performance Analysis and System Optimization,'' in \textit{IEEE Transactions on Communications}, vol. 71, no. 6, pp. 3387-3401, June 2023, doi: 10.1109/TCOMM.2023.3263566.
\bibitem{new4} “About INCOSE,” INCOSE CMS. https://www.incose.org/about-incose
\bibitem{new5} [1]“Project Performance International (PPI) | Systems Engineering Specialists,” PPI, Oct. 27, 2025. https://www.ppi-int.com/ (accessed Nov. 12, 2025).‌
\bibitem{b21} 3GPP, 3rd Generation Partnership Project (3GPP), Technical Report (TR) 38.901, 2024, version 18.0.0 Release 18.
\bibitem{b22} 3GPP, 3rd Generation Partnership Project (3GPP), Technical Report (TR) 38.843, 2023, version 18.0.0 Release 18.
\bibitem{b23} B. McMahan, E. Moore, D. Ramage, S. Hampson, and B.A. Arcas, ``Communication-efficient learning of deep networks from decentralized data``, Artificial Intelligence and Statistics (AISTAS), pp. 1273–1282, 2017.



\end{thebibliography}
\end{document}